# Generalized Biwords for Bitext Compression and Translation Spotting

**Felipe Sánchez-Martínez**                    FSANCHEZ@DLSI.UA.ES
**Rafael C. Carrasco**                          CARRASCO@DLSI.UA.ES
*Departament de Llenguatges i Sistemes Informàtics*
*Universitat d'Alacant, E-03071, Alacant, Spain*

**Miguel A. Martínez-Prieto**                   MIGUMAR2@INFOR.UVA.ES
**Joaquín Adiego**                              JADIEGO@INFOR.UVA.ES
*Departamento de Informática*
*Universidad de Valladolid, E-47011, Valladolid, Spain*

## Abstract

Large bilingual parallel texts (also known as *bitexts*) are usually stored in a compressed form, and previous work has shown that they can be more efficiently compressed if the fact that the two texts are mutual translations is exploited. For example, a bitext can be seen as a sequence of *biwords* —pairs of parallel words with a high probability of co-occurrence— that can be used as an intermediate representation in the compression process. However, the simple biword approach described in the literature can only exploit one-to-one word alignments and cannot tackle the reordering of words. We therefore introduce a generalization of biwords which can describe multi-word expressions and reorderings. We also describe some methods for the binary compression of generalized biword sequences, and compare their performance when different schemes are applied to the extraction of the biword sequence. In addition, we show that this generalization of biwords allows for the implementation of an efficient algorithm to look on the compressed bitext for words or text segments in one of the texts and retrieve their counterpart translations in the other text —an application usually referred to as *translation spotting*— with only some minor modifications in the compression algorithm.

## 1. Introduction

The increasing availability of large collections of multilingual texts has fostered the development of natural-language processing applications that address multilingual tasks —such as corpus-based machine translation (Arnold, Balkan, Meijer, Humphreys, & Sadler, 1994; Lopez, 2008; Koehn, 2010; Carl & Way, 2003), cross-language information retrieval (Grossman & Frieder, 2004, Ch. 4), the automatic extraction of bilingual lexicons (Tufis, Barbu, & Ion, 2004), and translation spotting (Simard, 2003; Véronis & Langlais, 2000). Other applications, which are monolingual in nature —e.g., syntactic parsing (Carroll, 2003), or word sense disambiguation (Ide & Véronis, 1998)— can also exploit multilingual texts by projecting the linguistic knowledge available in one language into other languages (Mihalcea & Simard, 2005).

A *bilingual parallel corpus*, or *bitext*, is a textual collection that contains pairs of documents which are translations of one another. The documents in a pair of this nature are sometimes called the *source text* and the *target text*, respectively. However, whenever the information as to how the document was created is unknown or irrelevant, the documents are simply called the *left text* and *right text*. In the words of Melamed (2001, p. 1), "bitexts are one of the richest sources of linguistic knowledge because the translation of a text into another language can be viewed as a detailed annotation of what that text means".

Bitexts are usually available in a compressed form in order to reduce storage requirements, to improve access times (Ziviani, Moura, Navarro, & Baeza-Yates, 2000), and to increase the efficiency of transmissions. However, the independent compression of the two texts of a bitext is clearly far from efficient because the information contained in both texts is redundant: in theory, one of the texts might be sufficient to generate a translated version if reliable machine translation systems were already available (Nevill-Manning & Bell, 1992). The improvement in compression performance obtained when taking advantage of the fact that the two texts in a bitext are mutual translations may be regarded as an indication of the quality of word alignments (Och & Ney, 2003). This indicator, which bounds the mutual information (Cover & Thomas, 1991) in the two texts of a bitext, does not require a manually-annotated corpus to evaluate the automatic alignment.

The first article dealing with the compression of bitexts was published by Nevill-Manning and Bell (1992). This approach compressed one of the texts in isolation, while the other was compressed by a general *prediction by partial matching* (PPM; Cleary & Witten, 1984) encoder based on a model that used the automatic translation of the left text to predict the words in the right text. This model exploited two types of relations —exact word matches and synonymy relationships provided by a thesaurus— and the relative weight of both predictions depended on the number of letters in the word that had been processed. This approach obtained better compression ratios than a standard PPM coder operating on the concatenated texts.

In contrast, Conley and Klein (2008) have proposed *text alignment* —that is, pairings between the words and phrases in one text and those in the other—, as the basis for multilingual text compression. Their algorithm extends the ideas of delta-encoding (Suel & Memon, 2003) to the case in which the right text $R$ is a translated, automatically aligned, version of the source text $L$: $L$ is compressed first, and each block in $R$ is then encoded as a reference to the parallel block in $L$. This method requires the computation of word- and phrase-level alignments, together with the lemmatized forms of $L$ and $R$. The translated text is retrieved from these references, using a bilingual glossary together with other linguistic resources: a lemmata dictionary of words in $L$, a dictionary with all the possible morphological variants of each word in $R$, and a bilingual glossary. The authors report slight improvements in the compression of the right text $R$ in comparison to classical compression algorithms such as BZIP2[1] or word-based Huffman (Moffat, 1989) (approximately 1% and 6%, respectively). However, the authors do not take into consideration the size of the auxiliary files needed for the retrieval of the right text.

In contrast to PPM, some text-compression methods use words rather than characters as input tokens (Moffat, 1989; Moffat & Isal, 2005). Analogously, Martínez-Prieto, Adiego, Sánchez-Martínez, de la Fuente, and Carrasco (2009), and Adiego and his colleagues (2009, 2010) propose the use of *biwords* —pairs of words, each one from a different text, with a high probability of co-occurrence— as input units for the compression of bitexts. This means that a biword-based intermediate representation of the bitext is obtained by exploiting alignments, and encoding unaligned words as pairs in which one component is the empty string. Significant spatial savings are achieved with this technique (Martínez-Prieto et al., 2009), although the compression of biword sequences requires larger dictionaries than the traditional text compression methods.

---

[1]. http://www.bzip.org

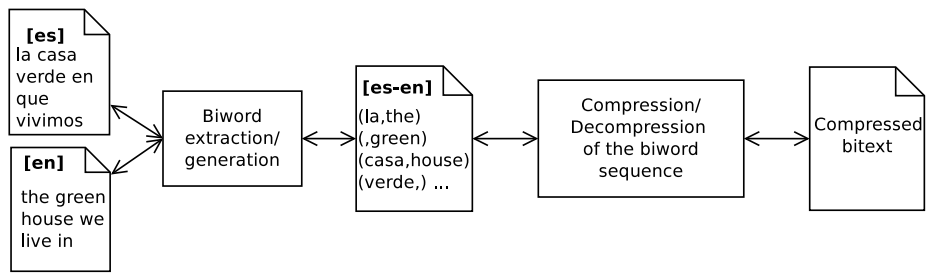

**Figure 1:** Processing pipeline of a biword-based bitext compression approach.

The biword-based compression approach works as a simple processing pipeline consisting of two stages (see Figure 1). After a text alignment has been obtained without pre-existing linguistic resources, the first stage transforms the bitext into a biword sequence. The second stage then compresses this sequence. Decompression works in reverse order: the biword sequence representing the bitext is first generated from the compressed file, and the original texts are then restored from this sequence.

A variation of the PPM algorithm that takes words rather than characters as input tokens and bytes rather than bits as minimal output units (Adiego & de la Fuente, 2006) can be directly applied in order to compress biword sequences. The bitext is thus compressed using a single probabilistic model for both texts —rather than the independent models used by older bitext-compression approaches (Nevill-Manning & Bell, 1992; Conley & Klein, 2008). The improvement over general-purpose compressors obtained with this approach depends on the language pair: for instance, a reduction in the output size of almost 11% is obtained for Spanish–Portuguese, and of about 2.5% for English–French (Martínez-Prieto et al., 2009).

A different biword-based scheme called 2LCAB has recently been proposed (Adiego et al., 2009) which creates a two-level dictionary to store the biwords and compresses the biword sequence with *End-Tagged Dense Code* (ETDC; Brisaboa, Fariña, Navarro, & Paramá, 2007). The usage of ETDC permits both Boyer-Moore-type searching (Boyer & Moore, 1977), and random access to the compressed file. If 2LCAB is used as a compression booster for a standard PPM coder, further improvements in compression are obtained, but it is no longer possible to directly search in the compressed files (Adiego et al., 2010).

The biword sequences obtained with the former biword-based compression methods contain a large fraction —between 10% and 60%, depending on the language pair— of *unpaired words*, that is, biwords of which one of the words in the pair is the empty word $\epsilon$. The unpaired words are generated in three different cases:

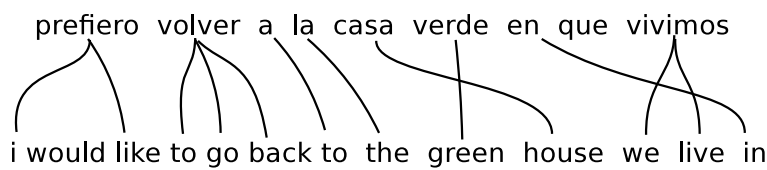

**Figure 2:** Example of a Spanish–English pair of sentences with one-to-many word alignments.

- The aligner is unable to connect a word with any of the words in the parallel text because, for example, infrequent idiomatic expressions or free translations have been found.

- The aligner generates a one-to-many alignment because a word has been translated into a multiword expression. For instance, if the Spanish word *volver* is translated into English as *to go back*, the biword extractor has to select one of the links, build a pair of words from that link, and leave the other words unpaired.

- The aligner generates some crossing alignments as a result of word reordering in the translation. For instance, in Figure 2, either the pair (verde, green) or the pair (casa, house) must be ignored by the biword extractor, thus leaving two unpaired words; otherwise, the information provided by the sequence will not be sufficient to retrieve both texts in the original order.

The last two sources of unpaired words are responsible for the different spatial savings reported by Martínez-Prieto et al. (2009) for bitexts consisting of closely-related languages (e.g., Spanish and Portuguese) and for those involving divergent language pairs (e.g., French and English), in which word reorderings and multiword translations are frequent.

In this paper, we describe and evaluate the simple biword extraction approach, and compare it with other schemes used to generate generalized biword sequences that maintain all or part of the structural information provided by the aligner. A biword essentially becomes a left word connected with a variable number of right words plus additional information concerning the relative position of each right word with regard to the preceding one. The fraction of unpaired words is thus reduced, and better compression ratios can be obtained.

We also show that this generalization of biwords allows for the implementation of an efficient *translation spotting* (Simard, 2003; Véronis & Langlais, 2000) algorithm on the compressed bitext; a task that consists of identifying the words (or text segments) in the other text that are the translation of the words in the query. Indeed, generalized biword sequences contain all the information needed in order to retrieve connected passages.

Generalized biwords can also be used as an ingredient in the bilingual language model employed in some statistical machine translation systems (Koehn, 2010). For instance, Mariño et al. (2006) use bilingual *n*-grams and consider the translation as a bilingual decoding process. Casacuberta and Vidal (2004) also exploit bilingual *n*-grams but apply stochastic finite-state transducers to this task. In both cases, the local reordering of words is addressed by considering multiword segments of source and target words as the fundamental translation units. Some alternative approaches (Niehues, Herrmann, Vogel, & Waibel, 2011; Matusov, Zens, Vilar, Mauser, Popović, Hasan, & Ney, 2006; Hasan, Ganitkevitch,

Ney, & Andrés-Ferrer, 2008) integrate bilingual language models as an additional feature in the decoding function that drives the statistical translation process. However, none of the approaches mentioned includes the structural information provided by the aligners as part of the bilingual language model.

The remainder of the paper is organized as follows. The following section shows how a generalized biword sequence can represent a bitext. Section 3 describes two different methods that can be applied to compress a biword sequence. Section 4 introduces the resources used to evaluate different generalizations of the biwords, whereas Section 5 discusses the compression results obtained. Section 6 then describes some modifications to one of these compression techniques in order to allow the compressed bitext to be searched and presents efficient search and translation spotting algorithms. Finally, some concluding remarks are presented in Section 7.

## 2. Extraction of Biword Sequences

Before extracting the sequence of biwords representing a bitext, the alignments between the words in the left text $L = l_1 l_2 \cdots l_M$ and the words in the right text $R = r_1 r_2 \cdots r_N$ must be established by the *word aligner*. Word aligners usually work after a *sentence aligner* has identified which pairs of sentences in the bitext are parallel, that is, a plausible mutual translation. Sentence alignment algorithms are often based on simple statistical models for the correlation between sentence lengths (Brown, Lai, & Mercer, 1991; Gale & Church, 1993).

Current word aligners use word-based statistical machine translation models (Brown, Cocke, Pietra, Pietra, Jelinek, Lafferty, Mercer, & Roossin, 1990; Brown, Pietra, Pietra, & Mercer, 1993) to compute the most likely alignment between the words in two parallel sentences (Koehn, 2010, Ch. 4). In our case, word alignments are computed with the open-source GIZA++ toolkit[2] (Och & Ney, 2003) which implements a set of methods, including standard word-based statistical machine translation models (Brown et al., 1993) and a hidden-Markov-model-based alignment model (Vogel, Ney, & Tillmann, 1996). GIZA++ produces alignments such as those depicted in Figure 2, in which a source word (here, a left word) can be aligned with many target words (here, right words), whereas a target word is aligned with, at most, one source word.

The result of *word alignment* is a bigraph $G = \{L, R, A\}$ in which an edge $\{l_i, r_j\} \in A$ between word $l_i \in L$ and word $r_j \in R$ signifies that they are mutual translations according to the translation model used by the aligner. These complex structures are processed by splitting the bigraph into connected components: each connected component is either an unpaired (right or left) word, or a left word $\sigma$ aligned with a sequence $\rho$ of (one or more) right words. As will be shown later, a connected component including the structural information needed to place all the words in their original positions in the bitext is what we term as a *generalized biword*.

In order to build a sequence $\mathcal{B}$ of generalized biwords, biwords will be sorted primarily according to their left component $\sigma$ and, secondarily, by the head of their right component $\rho$. More precisely, if left($\beta$) and right($\beta$) denote the left word and the sequence of right

---

words respectively in a biword $\beta$, and $\epsilon$ represents the empty left or right component in unpaired words, then $\alpha$ precedes $\beta$ if and only if:

- left($\alpha$) $\neq$ $\epsilon$, left($\beta$) $\neq$ $\epsilon$ and left($\alpha$) precedes left($\beta$) in $L$.

- Either left($\alpha$) = $\epsilon$ or left($\beta$) = $\epsilon$, and right($\alpha$) $\neq$ $\epsilon$, right($\beta$) $\neq$ $\epsilon$, and the initial word in right($\alpha$) precedes, in $R$, the initial word in right($\beta$).

- left($\alpha$) = $\epsilon$, left($\beta$) $\neq$ $\epsilon$ and there is no biword $\gamma$ such that $\beta$ precedes $\gamma$ and $\gamma$ precedes $\alpha$.

Every *generalized biword* $\beta = (\sigma, \rho, \omega)$ in the sequence $\mathcal{B}$ consists of:

- a string $\sigma$ in $\Sigma_L$,

- an array of strings $\rho$ in $\Sigma_R$, and

- an integer array $\omega$ containing one *offset* for every string in $\rho$.

Here, $\Sigma_L$ denotes the set of different words in $L$ enhanced with the empty word $\epsilon$, while $\Sigma_R$ denotes the set of subsequences in $R$ and includes the empty subsequence, represented by (). The array of offsets $\omega$ stores the structural information needed to place each word in $\rho$ in its original position.

The *offset* is a non-negative integer that specifies, for every word in $\rho$ that is not the first one, the number of words in $R$ located between this word and the preceding one in $\rho$, thus allowing the generation stage during decompression to keep track of the gaps in the subsequence $\rho$ that will be filled in with a word from a posterior biword in $\mathcal{B}$. The offset of the first word $w$ in $\rho \neq ()$ is defined as the number of words in $R$ located between $w$ and the first available gap, that is, the first word in $R$ that belongs to a biword that does not precede $\beta = (\sigma, \rho, \omega)$.

The combination of both types of offsets permits the encoding of translations with word reordering. Indeed, as can be seen in Figure 3, the offset in the biword (`casa`, (`house`),(`1`)) signifies that there is a one-word gap between `house` and `the` which is occupied by the word `green` with offset 0 in (`verde`,(`green`),(`0`)). The offsets in (`vivimos`, (`we,live`), (`0,0`)) indicate that `we` comes directly after the word `house` and `live` comes immediately after `we`. The pseudo-code of the procedure that extracts the sequence of generalized biwords and further details on its implementation can be found in Appendix A.

Henceforth, we shall call *biwords with shifts* those biwords with at least one non-null offset (*biwords without shifts*, otherwise). We shall further differentiate between *biwords with simple shifts*, where only the first offset is non-null, and *biwords with complex shifts*, with non-consecutive words in $R$.

Generalized biwords are clearly more expressive, and the bitexts will therefore be mapped onto shorter sequences. However, the enhanced variety of biwords implies that the compression algorithms must use larger dictionaries. The global effect on the compression ratio must therefore be explored. It is also worth measuring the effect of ignoring certain infrequent alignments in order to avoid biwords with complex shifts. For example, the generalized biword sequence in Figure 3 contains one biword with complex shifts, (`prefiero`,(`i,like`),(`0,1`)) which can be split into smaller components, such as ($\epsilon$,(`i`),

```
(prefiero,(i,like),(0,1))
(ε,(would),(0))
(volver,(to,go,back),(0,0,0))
(a,(to),(0))
(la,(the),(0))
(casa,(house),(1))
(verde, (green),(0))
(en,(in),(2))
(que,(),())
(vivimos,(we,live),(0,0))
```

**Figure 3:** Generalized biword sequence for the word-aligned sentence shown in Figure 2.

(0)) and (prefiero,(like),(0)), so that the sequence only includes biwords with simple shifts. If only simple shifts are allowed, the compression algorithm needs to encode, at most, one non-null offset per biword.

In the experiments we shall compare the results obtained when the algorithms described in the next section (Tre and 2lcab) are used in combination with four different methods to extract a sequence of biwords:

- **1:N Complex:** the one-to-many word alignments generated by Giza++ are used to generate a sequence of generalized biwords.

- **1:N Simple:** the biwords with complex shifts generated by the one-to-many alignments provided by Giza++ are split into biwords with simple shifts plus unpaired words; the result is a sequence of biwords with simple shifts or without shifts. Biwords with complex shifts are split by ignoring the least frequent alignments so that the resulting biwords only contain simple shifts.

- **1:1 Non-monotonic:** one-to-one word alignments are obtained by computing the intersection of the alignments produced by Giza++ when the left and the right text are exchanged; the result is a sequence of biwords whose right component contains, at most, one word (and these biwords cannot, therefore, have complex shifts).

- **1:1 Monotonic:** the 1:1 non-monotonic sequence is transformed into a sequence of biwords without shifts by splitting biwords with shifts into unpaired words.

The last method, **1:1 Monotonic**, does not use the enhancement provided by the generalization of biwords (i.e., the structural information), and is therefore equivalent to the basic procedures described earlier (Martínez-Prieto et al., 2009; Adiego et al., 2009, 2010).

## 3. Compression of Biword Sequences

It is clearly possible to compress the intermediate representation introduced in the previous section via the application of a wide range of approaches. Here, we describe and evaluate two different encoding methods, namely Tre (Subsection 3.2) and 2lcab (Subsection 3.3), that apply a word-based implementation of Huffman coding (Moffat & Turpin, 1997; Turpin

& Moffat, 2000) in which the input strings are mapped onto integers and then compressed with Huffman codewords (Huffman, 1952). Both methods encode the offsets as described below (Subsection 3.1) but differ in how they encode the lexical components of the biword sequence.

The use of Huffman codewords allows the two methods described here to achieve large spatial savings, but makes it inefficient to search in the compressed bitext and to retrieve matches. In Section 6 we shall describe a variant of the 2LCAB compression algorithm (searchable 2LCAB) which allows a sequence of words to be retrieved in the left text in addition to the parallel sequences (and their context) in the right text.

## 3.1 Structural Encoding

Preliminary tests showed that the biword extraction algorithm was capable of generating sequences with a high number (usually above 70%) of biwords without shifts, and an array of null values can thus be considered as the default offset sequence. The offsets can therefore be encoded as two streams of integer values:

- The positions $\mathcal{P} = (p_1, p_2, \ldots, p_N)$ of the biwords with shifts in the sequence $\mathcal{B}$. These positions can be expressed in relation to the previous biword with shifts in $\mathcal{B}$; for example, the biwords with shifts in Figure 3 are at $\mathcal{P} = (1, 5, 2)$.

- The offset values $\mathcal{O}$ for the biwords with shifts. In the example, the offsets are $\mathcal{O} = (0, 1, 1, 2)$; the first two offsets belong to the first biword with shifts whereas the following ones belong to the second and third biword with shifts, respectively.

Both streams are therefore encoded by using two independent sets of Huffman codewords.

## 3.2 The TRE Compressor

The *Translation Relationship-based Encoder* (TRE) assigns codewords to the left word and to the sequences of right words in the biword through the use of two independent methods. The left text is encoded using word-based Huffman coding (Moffat, 1989). In contrast, the right text is encoded by using the left text as its context. To do this, TRE uses three dictionaries: one, $\Sigma_L$, with the left words, a second one, $\Sigma_R$, with the sequences of right words, and a third one, the *translation dictionary* $\tau_{\mathcal{B}}$, which maps each word $\sigma \in \Sigma_L$ onto a subset of entries in $\Sigma_R$:

$$\tau_{\mathcal{B}}(\sigma) = \{\rho \in \Sigma_R : \exists \omega \in \mathbb{N}^* : (\sigma, \rho, \omega) \in \mathcal{B}\}.$$

For every $\sigma \in \Sigma_L$ the sequences in $\tau_{\mathcal{B}}(\sigma)$ are sorted by frequency and assigned an integer in the range $[1, |\tau_{\mathcal{B}}(\sigma)|]$, thus signifying that the most frequent translations have the lowest values.

At the compression stage, the text in every biword $(\sigma, \rho, \omega)$ is mapped onto a pair of integers —a reference to the left word $\sigma$ and the integer value assigned to the sequence of right words $\rho$ by $\tau_{\mathcal{B}}(\sigma)$—, and both sequences of integers are then compressed using independent Huffman codewords. The compression efficiency is improved because the most frequent translations are all assigned low (and thus recurrent) integer values. Finally, the compressed file includes a header with:

- The dictionaries $\Sigma_L$ and $\Sigma_R$, which are independently encoded using PPM compression. A special character is used to separate consecutive entries in the dictionaries and white-space serves as the delimiter in word sequences.

- The translation dictionary $\tau_\mathcal{B}$, that is, a Huffman-compressed sequence of integers (Moffat & Turpin, 1997) containing, for every entry $\sigma \in \Sigma_L$, the size of $\tau_\mathcal{B}(\sigma)$ and the references to the entries in $\Sigma_R$ that store every sequence in $\tau_\mathcal{B}(\sigma)$.

- The independent Huffman codewords used to compress the integer sequences of $\sigma$-references, and $\tau_\mathcal{B}(\sigma)$ values.

### 3.3 The 2LCAB Compressor

In contrast to TRE, the 2-Level Compressor for Aligned Bitexts (2LCAB; Adiego et al., 2009) encodes every biword with a single codeword based on a two-level dictionary. The first level consists of two dictionaries, $\Sigma_L$ and $\Sigma_R$, containing the left words and the sequences of right words, respectively, that appear in the biword sequence $\mathcal{B}$. The second level dictionary $\Sigma_B$ stores the different biwords in $\mathcal{B}$ as an integer sequence of alternating references to the entries in $\Sigma_L$ and $\Sigma_R$. The text in the sequence $\mathcal{B}$ can then be mapped onto a sequence of references to entries in $\Sigma_B$.

The header includes $\Sigma_L$ and $\Sigma_R$ which are compressed, as in TRE, with a PPM algorithm (Cleary & Witten, 1984). It also contains the codewords (selected according to the Huffman compression procedure) for the integers in the sequence describing the dictionary $\Sigma_B$, the encoded dictionary $\Sigma_B$, and a second list of Huffman codewords used to encode the biword sequence $\mathcal{B}$. This implementation of 2LCAB employs (bit-oriented) Huffman coding, but the original work (Adiego et al., 2009), and the application described in Section 6 implement byte-oriented ETDC (Brisaboa et al., 2007). The bit-oriented approach is more effective, but ETDC permits faster searches on the compressed bitext.

## 4. Resources and Settings

In order to evaluate the performance of the bitext compressors based on generalized biwords we have made use of the following bitext collections:

- A 100 MB Spanish–Catalan (`es-ca`) bitext obtained from *El Periódico de Catalunya*,[3] a daily newspaper published in Catalan and Spanish.

- A 100 MB Welsh–English (`cy-en`) bitext from the *Proceedings of the National Assembly for Wales* (Jones & Eisele, 2006).[4]

- Bitexts (100 MB each) from the *European Parliament Proceedings Parallel Corpus* (Europarl; Koehn, 2005) for seven different language pairs: German–English (`de-en`), Spanish–English (`es-en`), Spanish–French (`es-fr`), Spanish–Italian (`es-it`), Spanish–Portuguese (`es-pt`), French–English (`fr-en`), and Finnish–English (`fi-en`).

---

3. Available on-line at `http://www.elperiodico.com`.
4. Available on-line at `http://xixona.dlsi.ua.es/corpora/UAGT-PNAW/`.

| Lang. pair | BZIP2 | GZIP | P7ZIP | PPMDI | WH |
|---|---|---|---|---|---|
| en-de | 22.19% | 31.59% | 21.61% | **20.35%** | 23.28% |
| es-en | 22.01% | 31.38% | 21.39% | **20.12%** | 23.95% |
| fr-es | 21.51% | 30.80% | 20.75% | **19.57%** | 24.04% |
| it-es | 21.88% | 31.00% | 21.15% | **19.98%** | 23.57% |
| pt-es | 21.94% | 31.10% | 21.11% | **20.02%** | 23.11% |
| fr-en | 21.65% | 31.10% | 21.06% | **19.76%** | 24.57% |
| es-ca | 27.21% | 37.09% | **24.41%** | 25.43% | 27.66% |
| en-fi | 22.17% | 31.32% | 20.78% | **20.33%** | 22.97% |
| cy-en | 21.83% | 31.11% | 20.33% | **20.07%** | 25.18% |

**Table 1:** Compression ratios obtained with four general-purpose compressors and a word-based text compressor (WH).

As is common in information retrieval applications, the texts were tokenized and converted to lowercase (Manning & Schütze, 1999, Ch. 4). The tokenization placed blank spaces before and after every punctuation mark, and a word was thus defined as being any sequence of alphanumeric characters delimited by blank spaces.

Word alignments were computed with the Giza++ toolkit, with all parameters set to their default values, with the exception of the fertility which was set to 5 (the default being 9). The *fertility* is the maximum number of words with which a word can be aligned, and a low value moderates the number of right sequences with one single occurrence in the bitext.

## 5. Results and Discussion

As a reference, Table 1 shows the compression ratio —defined as the quotient between the lengths of the output and the input texts (Ziv & Lempel, 1977)— achieved with the aforementioned bitexts when they are compressed with a variety of general-purpose compressors and with a word-based compressor operating on the concatenation of the two texts $L$ and $R$. The other approaches quoted in the introduction could not be compared because either the code or the linguistic resources required were not publicly available. The compressors used as a reference are:

- The BZIP2 compressor,[5] which splits the text into blocks (100-900 KB), then, applies the *Burrows-Wheeler Transform* (BWT; Burrows & Wheeler, 1994) followed by a *move-to-front* transformation and, finally, encodes the result with a Huffman encoder.

- Two dictionary-based compressors built on different variants of the Ziv-Lempel's LZ77 (Ziv & Lempel, 1977) algorithm. First, GZIP,[6] a classical compressor that combines LZ77-based modeling with Huffman coding (Huffman, 1952). Second, the modern P7ZIP[7] compressor based on the *Lempel-Ziv-Markov chain algorithm* (Sa-lomon, 2007, Sec. 3.24), an algorithm which improves LZ77 with a large dictionary (up to 4 GB) and range encoding (Martin, 1979).

---

5. http://www.bzip.org. Experiments run with version 1.0.5.
6. http://www.gzip.org. Experiments were carried out with version 1.3.12-6.
7. http://www.7zip.com. Experiments were carried out using version 4.58~dfsg1.1.

| Lang. pair | 1:1 Monotonic | 1:1 Non-monotonic | 1:N Simple | 1:N Complex |
|---|---|---|---|---|
| en-de | 20.38% | **20.06%** | 20.26% | 21.22% |
| es-en | 19.63% | 18.85% | **18.69%** | 19.33% |
| fr-es | 19.07% | **18.60%** | 18.78% | 19.51% |
| it-es | 19.21% | **18.86%** | 19.11% | 20.00% |
| pt-es | 18.44% | **18.06%** | 18.17% | 18.79% |
| fr-en | 20.20% | **19.30%** | 19.31% | 20.06% |
| es-ca | 17.02% | 16.95% | **16.78%** | 16.86% |
| en-fi | 21.50% | **20.82%** | 21.70% | 22.24% |
| cy-en | 20.06% | 18.69% | **18.05%** | 18.22% |

**Table 2:** Compression ratios obtained with the TRE compressor and different biword extraction methods.

- PPMDI (Shkarin, 2002) as a representative of the *Prediction by Partial Matching* (PPM; Cleary & Witten, 1984) family of compressors. PPMDI uses a high-order context model and method D (Howard & Vitter, 1992) to handle escape codes. The implementation available in the *Pizza&Chili* website[8] with the default configuration (sixth-order context model) has been used.

- A word-based Huffman compressor (Moffat, 1989) that maps the input strings to integers before encoding the values with Huffman codewords (Moffat & Turpin, 1997; Turpin & Moffat, 2000). This method was originally designed to compress text, but also works well with other types of sources. The dictionary that maps words to integers —after its encoding with a PPMDI compressor— is part of the output.

As can be seen in Table 1, the lowest compression ratios are obtained with PPMDI, except for the es-ca pair. The fact that the compression ratios depend only moderately on the languages involved suggests that these compressors do not benefit from the (variable) cross-language information provided by the translations.

These ratios must be compared with the performance of the two compressors described in this section and presented in Tables 2 and 3. Note that although all the bitexts were aligned in both translation directions, only the results obtained with the direction producing the best compression are reported here, since the effect of this choice on the compression ratio proved to be small (an average difference of 0.2 percentage points).

The comparison shows that both TRE and 2LCAB outperform the general-purpose compressors in all cases but that of the en-fi pair. The best results are obtained in most of the cases when one-to-one alignments are used with both techniques. 2LCAB achieves slightly better results than TRE for all language pairs with the exception of it-es and pt-es, although in these two cases the difference in performance is too small to be considered relevant. The low performance for en-fi is the consequence of the larger translation dictionaries used by TRE, and the larger bilingual dictionary used by 2LCAB, in comparison to the other language pairs. Furthermore, the percentage of unpaired words is also higher than that of the other language pairs as will be seen below.

---

8. http://pizzachili.dcc.uchile.cl/utils/ppmdi.tar.gz

| Lang. pair | 1:1 Monotonic | 1:1 Non-monotonic | 1:N Simple | 1:N Complex |
|---|---|---|---|---|
| en-de | 19.98% | **19.83%** | 20.77% | 22.12% |
| es-en | 19.29% | **18.68%** | 19.08% | 19.99% |
| fr-es | 18.89% | **18.50%** | 19.27% | 20.25% |
| it-es | 19.18% | **18.87%** | 19.77% | 20.96% |
| pt-es | 18.46% | **18.09%** | 18.75% | 19.60% |
| fr-en | 19.75% | **19.03%** | 19.65% | 20.71% |
| es-ca | 16.69% | 16.61% | **16.59%** | 16.70% |
| en-fi | 21.29% | **20.62%** | 22.46% | 23.31% |
| cy-en | 19.43% | 18.30% | **17.98%** | 18.25% |

**Table 3:** Compression ratios obtained with the 2LCAB compressor and different biword extraction methods.

| Lang. pair | General purpose | 2LCAB 1:1 Monotonic | 2LCAB (best) | Gain (Best/Gen.) | Gain (Best/Mono.) |
|---|---|---|---|---|---|
| en-de | 20.35% | 19.98% | 19.83% | 2.56% | 0.75% |
| es-en | 20.12% | 19.29% | 18.68% | 7.16% | 3.16% |
| fr-es | 19.57% | 18.89% | 18.50% | 5.47% | 2.06% |
| it-es | 19.98% | 19.18% | 18.87% | 5.56% | 1.62% |
| pt-es | 20.02% | 18.46% | 18.09% | 9.64% | 2.00% |
| fr-en | 19.76% | 19.75% | 19.03% | 3.69% | 3.65% |
| es-ca | 24.41% | 16.69% | 16.59% | 32.04% | 0.60% |
| en-fi | 20.33% | 21.29% | 20.62% | -1.43% | 3.15% |
| cy-en | 20.07% | 19.43% | 17.98% | 10.41% | 7.46% |

**Table 4:** Summary of the best compression results obtained with: i) general-purpose and word-based compressors; ii) the 2LCAB compressor with no structural information (1:1 Monotonic); iii) the best 2LCAB compressor. The columns on the right show the relative improvement of the best 2LCAB over the general purpose and monotonic compressors, respectively.

Table 4 summarizes the results obtained by the general-purpose compressors and by 2LCAB, and the relative gains in compression performance with regard to the general-purpose compressors performing best, and with regard to that of 2LCAB when compressing the 1:1 Monotonic biword sequence. The greatest improvement, in comparison to the results obtained for the general-purpose compressors, is achieved for the language pair es-ca: instead of the 24.4% compression ratio obtained by P7ZIP, 2LCAB achieves a compression ratio of 16.6% which represents a substantial spatial saving (32.04% relative improvement). This suggests that TRE and 2LCAB take advantage of the fact that the texts contain the same information but "encoded" in different languages, particularly in the case of highly parallel bitexts (`en-cy`) and languages with a high syntactic correlation (`es-ca`).

| Lang. pair | 1:1 Monotonic | 1:1 Non-monotonic | 1:N Simple | 1:N Complex |
|---|---|---|---|---|
| en-de | 0.600 | 0.483 | 0.352 | 0.319 |
| es-en | 0.506 | 0.395 | 0.984 | 0.258 |
| fr-es | 0.463 | 0.398 | 0.290 | 0.266 |
| it-es | 0.465 | 0.408 | 0.279 | 0.247 |
| pt-es | 0.421 | 0.363 | 0.249 | 0.225 |
| fr-en | 0.540 | 0.425 | 0.316 | 0.292 |
| es-ca | 0.128 | 0.122 | 0.077 | 0.072 |
| en-fi | 0.684 | 0.610 | 0.492 | 0.467 |
| cy-en | 0.530 | 0.387 | 0.286 | 0.276 |

**Table 5:** Fraction of biwords in the extracted sequence with one empty component.

The generalization of biwords generates shorter biword sequences, essentially because the sequence extracted contains a lower fraction of unpaired words. Table 5 shows the fraction of biwords in which one component is the empty word (the other being an unpaired word). This number is obviously considerably reduced when offsets are used to encode the structural information implicit in the alignments. Of course, in the case of the `1:N Complex` approach, the fraction coincides with that of the bigraph produced by the aligner.

As expected, the effect of the generalization on the percentage of biwords with an empty component depends on the languages involved, the reduction being smaller for pairs of closely-related languages (`es-ca`, `pt-es`, and `it-es`) than for pairs of languages with strong grammatical divergences (`en-de`, `es-en`, and `fr-en`) since, in the latter case, word reorderings and multiword expressions commonly appear in translations.

In order to gain some insight into the performance of the TRE and 2LCAB compressors, it is interesting to make a separate examination of the contribution to the output size of the headers, the dictionaries and the codewords. In this respect, it is worth noting that:

1. The number of entries in the left dictionary $\Sigma_L$ does not depend on the method used to extract the biword sequence.

2. The number of entries in the right dictionary $\Sigma_R$ is identical for the two extraction methods based on one-to-one alignments because it consists of all the words found in the right text plus the empty word.

Tables 6 and 7 show the fraction of the output that corresponds to the encoded biword sequence (columns B), and to the translation dictionary or biword dictionary (columns D), depending on the compressor used. These numbers reveal that the more general the biwords used are, the more compact the encoded sequences are compared to the headers and dictionaries. In particular, the translation dictionary size (analogously, the biword dictionary size) does not differ much if non-monotonic alignments are used instead of the basic method. However, the usage of one-to-many alignments causes the size of the dictionary to grow considerably, particularly in the case of the biword dictionary $\Sigma_B$ used by 2LCAB.

| Lang. pair | 1:1 Monotonic | | 1:1 Non-monotonic | | 1:N Simple | | 1:N Complex | |
|---|---|---|---|---|---|---|---|---|
| | B | D | B | D | B | D | B | D |
| en-de | 0.933 | 0.031 | 0.930 | 0.034 | 0.857 | 0.070 | 0.787 | 0.078 |
| es-en | 0.942 | 0.035 | 0.940 | 0.037 | 0.894 | 0.058 | 0.839 | 0.069 |
| fr-es | 0.934 | 0.041 | 0.932 | 0.042 | 0.884 | 0.065 | 0.826 | 0.074 |
| it-es | 0.928 | 0.045 | 0.927 | 0.046 | 0.870 | 0.073 | 0.805 | 0.081 |
| pt-es | 0.923 | 0.047 | 0.922 | 0.048 | 0.870 | 0.072 | 0.819 | 0.078 |
| fr-en | 0.951 | 0.029 | 0.948 | 0.031 | 0.903 | 0.054 | 0.845 | 0.066 |
| es-ca | 0.892 | 0.039 | 0.893 | 0.039 | 0.870 | 0.045 | 0.862 | 0.046 |
| en-fi | 0.884 | 0.059 | 0.884 | 0.058 | 0.785 | 0.093 | 0.724 | 0.089 |
| cy-en | 0.960 | 0.021 | 0.958 | 0.023 | 0.933 | 0.034 | 0.916 | 0.038 |

**Table 6:** Fraction of the file compressed with TRE encoding the bitext (B) and the translation dictionary (D).

| Lang. pair | 1:1 Monotonic | | 1:1 Non-monotonic | | 1:N Simple | | 1:N Complex | |
|---|---|---|---|---|---|---|---|---|
| | B | D | B | D | B | D | B | D |
| en-de | 0.896 | 0.065 | 0.893 | 0.067 | 0.808 | 0.116 | 0.733 | 0.130 |
| es-en | 0.911 | 0.063 | 0.908 | 0.065 | 0.844 | 0.105 | 0.788 | 0.119 |
| fr-es | 0.894 | 0.077 | 0.892 | 0.079 | 0.831 | 0.115 | 0.772 | 0.126 |
| it-es | 0.886 | 0.083 | 0.884 | 0.084 | 0.812 | 0.128 | 0.745 | 0.140 |
| pt-es | 0.880 | 0.087 | 0.880 | 0.088 | 0.814 | 0.126 | 0.762 | 0.133 |
| fr-en | 0.920 | 0.057 | 0.916 | 0.060 | 0.856 | 0.098 | 0.796 | 0.114 |
| es-ca | 0.859 | 0.071 | 0.860 | 0.070 | 0.832 | 0.080 | 0.824 | 0.082 |
| en-fi | 0.845 | 0.094 | 0.844 | 0.093 | 0.725 | 0.151 | 0.666 | 0.147 |
| cy-en | 0.939 | 0.039 | 0.935 | 0.042 | 0.903 | 0.061 | 0.884 | 0.066 |

**Table 7:** Fraction of the file compressed with 2LCAB encoding the bitext (B) and the biword dictionary (D).

| Lang. pair | 1:N Simple | | | 1:N Complex | | |
|---|---|---|---|---|---|---|
| | Comp. Ratio | Biword reduc. | $\tau_{\mathcal{B}}$ weight | Comp. Ratio | Biword reduc. | $\tau_{\mathcal{B}}$ weight |
| en-de | 19.82% | 0.514 | 0.033 | 19.93% | 0.465 | 0.033 |
| es-en | 18.51% | 0.495 | 0.024 | 18.56% | 0.441 | 0.024 |
| fr-es | 18.58% | 0.472 | 0.028 | 18.64% | 0.434 | 0.027 |
| it-es | 18.82% | 0.458 | 0.029 | 18.91% | 0.418 | 0.029 |
| pt-es | 17.98% | 0.483 | 0.030 | 18.05% | 0.456 | 0.030 |
| fr-en | 19.10% | 0.479 | 0.022 | 19.17% | 0.416 | 0.022 |
| es-ca | 16.78% | 1.000 | 0.045 | 16.86% | 1.000 | 0.046 |
| en-fi | 21.01% | 0.539 | 0.049 | 21.09% | 0.559 | 0.048 |
| cy-en | 18.05% | 1.000 | 0.034 | 18.08% | 0.631 | 0.019 |

**Table 8:** Compression ratios with TRE when infrequent biwords are split into smaller biwords, remaining fraction of the original biword dictionary, and relative size of the translation dictionary in the compressed file.

| Lang. pair | 1:N Simple | | | 1:N Complex | | |
|---|---|---|---|---|---|---|
| | Comp. Ratio | Biword reduc. | $\Sigma_{\mathcal{B}}$ weight | Comp. Ratio | Biword reduc. | $\Sigma_{\mathcal{B}}$ weight |
| en-de | 19.42% | 0.435 | 0.043 | 19.51% | 0.393 | 0.043 |
| es-en | 18.16% | 0.406 | 0.036 | 18.22% | 0.326 | 0.031 |
| fr-es | 18.21% | 0.385 | 0.038 | 18.27% | 0.354 | 0.037 |
| it-es | 18.47% | 0.334 | 0.035 | 18.55% | 0.306 | 0.035 |
| pt-es | 17.67% | 0.361 | 0.037 | 17.73% | 0.341 | 0.036 |
| fr-en | 18.71% | 0.386 | 0.033 | 18.77% | 0.334 | 0.032 |
| es-ca | 16.59% | 1.000 | 0.080 | 16.70% | 1.000 | 0.082 |
| en-fi | 20.50% | 0.445 | 0.056 | 20.55% | 0.464 | 0.055 |
| cy-en | 17.68% | 0.573 | 0.029 | 17.71% | 0.542 | 0.029 |

**Table 9:** Compression ratios with 2LCAB when infrequent biwords are split into smaller biwords, remaining fraction of the original biword dictionary, and relative size of the biword dictionary in the compressed file.

This, in most of the cases, causes 2LCAB to perform worse than TRE with one-to-many word alignments.

The observation that a one-to-many alignment leads to larger dictionaries makes it worth exploring the effect on the compression ratio when very infrequent biwords are discarded. Tables 8 and 9 therefore show the compression ratios obtained by TRE and 2LCAB, respectively, when infrequent biwords are split into smaller, more frequent, biwords. This split proceeds iteratively by removing the least frequent alignment in the biword (which produces a new unpaired word) until all biword frequencies are above a threshold $\delta$ or the biwords only contain unpaired words. The threshold is determined by means of a ternary search which optimizes the compression ratio. The tables also show the fraction of biwords that remain in the dictionary after pruning, along with the new fraction of the output that corresponds to the translation or biword dictionary.

As can be seen in the tables, discarding the most infrequent biwords (about two thirds of them) usually leads to an improvement in the compression ratios, except in the case of very similar languages, such as Catalan and Spanish, in which the translation is highly parallel. This effect is more important in the case of 2LCAB because the pruning leads to a large reduction in the size of the biword dictionary and this compensates the small increment in the total number of biwords needed to represent the bitext (between 5% and 10% of increment depending on the method used for its generation). With this filtering, 2LCAB and TRE obtain the best results when extracting the biword sequence with method `1:N Simple`.

## 6. Translation Spotting with Compressed Bitexts

The exploitation of bitexts by computer-aided translation tools has evolved from simple *bilingual concordancers* (Barlow, 2004; Bowker & Barlow, 2004) to advanced *translation search engines* (Callison-Burch, Bannard, & Schroeder, 2005a; Bourdaillet, Huet, Langlais, & Lapalme, 2010). The standard translation unit processed by bilingual concordancers are sentences, and these concordancers can thus only provide a whole sentence as the result of a translation search. In contrast, translation search engines have translation spotting capabilities, i.e. they can retrieve parallel text segments in bitexts.

It would seem that existing translation search engines (Callison-Burch et al., 2005a; Bourdaillet et al., 2010) do not access bitexts in their compressed forms because storing the correspondences between the translated segments requires additional data structures such as word indexes or suffix arrays (Lopez, 2007; Callison-Burch, Bannard, & Schroeder, 2005b); suffix arrays typically require four times the size of the text (Manber & Myers, 1993). In contrast, the generalized biwords require much less space, they integrate the alignment information into the compressed bitext, and this information can be exploited to retrieve translation examples. In this section we describe some minor modifications that need to be done to the 2LCAB compression algorithm before it can be applied to this task. We also describe a search algorithm on the compressed bitexts and evaluate the compression performance of the new 2LCAB implementation (searchable 2LCAB).

The application of the 2LCAB compression technique to the direct search in compressed bitexts leads to certain challenges which are not present during the decompression process because:

- Huffman and PPM compression hinder both direct searching and random access to the compressed text (Bell, Cleary, & Witten, 1990).

- In a multilevel scheme such as 2LCAB, whenever a matching string is found —for instance, in the biword dictionary $\Sigma_B$— it is necessary to know the string's codeword in order to search for the encoded string in the higher level —for example, the sequence of biwords $\mathcal{B}$.

The differences induced in 2LCAB are described as follows, and are summarized in Table 10.

| Data | Content | 2LCAB | Searchable 2LCAB |
|------|---------|-------|------------------|
| $\Sigma_L$ | lword0, lword1, . . . | PPM | PPM[†] |
| $\Sigma_R$ | rword0, rword1, . . . | PPM | PPM[†] |
| $\Sigma_B$ | lpos0, rpos0, lpos1, rpos1, . . . | Huffman | ETDC[†] |
| $\mathcal{B}$ | bpos0, bpos1, . . . | Huffman | ETDC |
| $\mathcal{P}$ | delta0, delta1, . . . | Huffman | RRR |
| $\mathcal{O}$ | offset0, offset1, . . . | Huffman | DAC & RG |

**Table 10:** Summary of the compression methods applied. Those marked with † sort the items before the compression. RG is only used with DAC in case biwords with complex shifts are present.

## 6.1 Searchable 2LCAB Compression

There are several alternative compression methods, such as ETDC, which allow direct searches in compressed text. In contrast to the output of the Huffman compression, the ETDC header only stores words, because each codeword can be derived from the word position —henceforth, its rank— if the words are sorted according to their relative frequencies in the document which is to be compressed. This means that there is always a mapping, denoted code($n$), which provides the ETDC codeword for the $n$-th most frequent word, along with the corresponding reverse mapping.

For instance, if the $b$-th byte in the compressed bilingual dictionary $\Sigma_B$ matches a left-word code, its rank in $\Sigma_B$ determines which codeword must be looked for in $\mathcal{B}$.[9] Of course, this rank can be obtained by keeping a record of the number $n$ of words in $\Sigma_B$ scanned so far, but the standard pattern matching algorithms —such as BM (Boyer & Moore, 1977) or KMP (Knuth, Morris, & Pratt, 1977)— are not well suited to this tracking. We therefore use a finite sequence of bits $S = b_1, b_2, \cdots, b_{|S|}$ to retrieve the rank of the $n$-th byte $\beta_n$ in the encoded $\Sigma_B$. This sequence has $b_n = 1$ for every $n$ such that $\beta_n$ is the final byte in a codeword and it can be built on the fly when $\Sigma_B$ is read from the compressed file.

*Succinct data structures* (Navarro & Mäkinen, 2007, Sec. 6), such as RG (González, Grabowski, Mäkinen, & Navarro, 2005) and RRR (Raman, Raman, & Rao, 2002), provide an effective way in which to represent a sequence of bits and recover the rank associated with every matching sequence, because they support a number of operations in the sequence of bits $S$ in a time that is independent of its length (Clark, 1996):

- The number of bits $\text{rank}_S(i)$ whose value is one in $b_1 \cdots b_i$.

- The position $\text{select}_S(i)$ of the $i$-th bit in $S$ whose value is one;

- The value of the $i$-th bit in $S$, denoted as $\text{access}_S(i) = b_i$.

Moreover, the structural information in $\mathcal{P}$ describing which the biwords with shifts are also needs to be randomly accessed, and the succinct data structure RRR provides a compact alternative to the Huffman-based method used in 2LCAB to compress $\mathcal{P}$, the sequence of position increments. Indeed, the RRR encoding is especially compact when the information is unbalanced —for instance, above 80% of the bits show identical value—, as in this case in which the number of biwords with shifts is small. Therefore, the integer sequence $\mathcal{P}$ will instead be stored as a binary sequence $P = p_1, p_2, \ldots, p_m$ such that $p_i = 1$ if the $n$-th byte in $\mathcal{B}$ is the final byte in the codeword of a biword with shifts.

---

9. This rank can also be used to discard false matches originated by a coincidence with a right word in $\Sigma_B$ because, in such cases, the rank will be an even number.

Finally, the offsets stored in $\mathcal{O}$ will be compressed with *directly addressable variable-length codes* (DAC, Brisaboa, Ladra, & Navarro, 2009) which, in contrast to the Huffman compression, provide direct access to the $n$-th encoded element. Information associated with the $n$-th biword can thus be retrieved immediately, since DAC encoding does not require the preceding sequence to be decompressed from the beginning. As the biwords with complex shifts contain more than one offset, the access to $\mathcal{O}$ in these cases is indirect and provided through an auxiliary RG data structure. This structure builds a sequence of bits $Q = q_1, q_2, \cdots, q_{|Q|}$ where $|Q|$ is the total number of offsets stored in $\mathcal{O}$. The sequence $Q$ has $q_i = 1$ if $\mathcal{O}_i$ is the first offset in the array $\omega$ of a biword $(\sigma, \rho, \omega)$, and $q_i = 0$ otherwise.

As can be seen in Table 10, the searchable 2LCAB method replaces the Huffman compression with ETDC and sorts some contents so that the higher-level ETDC compression does not need to store codewords in its header.

## 6.2 Translation Spotting

The searchable 2LCAB described above is complemented with a search algorithm which, given a single word $w$ in the left text, proceeds as follows:

1. The word $w$ is looked for in $\Sigma_L$ —whose relatively small size permits an uncompressed copy to be stored in the memory— and its identifier $n$, given by the word position in $\Sigma_L$, is used to obtain its ETDC codeword $c = \text{code}(n)$.

2. An exact pattern-matching algorithm (Knuth et al., 1977) identifies all the occurrences of the codeword $c$ in the biword dictionary $\Sigma_B$. If a match is found at the $b$-th byte and $r = \text{rank}_S(b)$ is odd (indicating a match with a $\Sigma_L$-codeword, that is, a biword with a left component $w$), then, the biword with codeword $\text{code}(r/2)$ is added to the search set $Z$.

3. The multi-pattern matching algorithm SET-HORSPOOL (Horspool, 1980; Navarro & Raffinot, 2002) locates all the codewords in the sequence of biwords $\mathcal{B}$ that match one of those contained in $Z$, and the matching positions are added to a new set $M$.

4. For every match $m \in M$, the adjacent right component is read from $\mathcal{B}$ and, whenever $p_m = 1$ in $P$, the offsets are recovered from $\mathcal{O}$ and used to place the right words in the original order. In case the biwords can have complex shifts, the interval $\mathcal{O}_i \cdots \mathcal{O}_j$ containing the offsets $\omega$ starts at $i = \text{select}_Q(r)$ and ends at $j = \text{select}_Q(r+1) - 1$, with $r = \text{rank}_P(m)$.

In case the query consists of a sequence of words $(w_1, w_2, \cdots w_K)$ with $K > 1$, the SET-HORSPOOL algorithm is executed only for the word $w_k$ in the sequence generating the smallest set of codewords to locate $Z_k$, and the remaining words are then used to filter the results once the biword context has been retrieved.

Table 11 shows an actual example of the output obtained for a multiple word query and a compressed biword sequence obtained with the `1:N Complex` method. Note that the third match shows a non-contiguous translation, a case which cannot be retrieved with the original 2LCAB implementation (Adiego et al., 2009).

| | |
|---|---|
| Left text: | it is only democratic that our citizens should be able to exercise influence , and it goes without saying that they should be entitled to all the information they need **in order to perform** their civic duties in society . |
| Right text: | un componente de la democracia es que los ciudadanos puedan influir y , obviamente , que tengan derecho a acceder a la información necesaria **para actuar** como ciudadanos en sus sociedades . |
| Left text: | it is concerned , then , with protection under criminal law and with europol units having to receive the information and intelligence they need **in order to perform** their tasks . |
| Right text: | se trata , por tanto , de protección penal y de que las unidades de europol deben obtener la información y los datos que necesiten **para poder realizar** su trabajo . |
| Left text: | the co-decision procedure must be used **in order to perform** this legislative work under conditions which guarantee a genuine debate , involving society and citizens . |
| Right text: | **para que** ese trabajo legislativo se **realice** en condiciones que garanticen un verdadero debate , social y ciudadano , hace falta recurrir al procedimiento de codecisión . |
| Left text: | what are the tools , what are the procedures , that we need **in order to perform** them ? |
| Right text: | ¿ cuáles son las herramientas , los procedimientos que necesitamos **para ejecutarlas** ? |

**Table 11:** Output obtained after the query "*in order to perform*" on the bitext compressed with the `1:N Complex` method. The query terms and their translations are spotted in boldface.

## 6.3 Experimental Evaluation

The compression ratios obtained with the searchable 2LCAB are shown in Table 12. The algorithm is clearly not as effective as the 2LCAB described in Section 3, leading to compression ratios which are slightly worse than those obtained with general purpose and word-based compressors. However, it is worth to mentioning that these compressed files include the information concerning the alignments between the words, information that is not included in the files compressed with the standard compressors but is necessary to perform translation spotting.

Table 12 also shows that the `1:1 Monotonic` method is in this case more effective than the `1:1 Non-monotonic` method because the latter needs an additional data structure (the RRR bit sequence) in order to access the structural information. Moreover, the byte orientation of ETDC reduces the gain obtained by encoding a lower number of biwords.

| Lang.<br>pair | 1:1<br>Monotonic | 1:1<br>Non-monotonic | 1:N<br>Simple | 1:N<br>Complex |
|---|---|---|---|---|
| en-de | **22.66%** | 23.20% | 24.10% | 26.02% |
| es-en | 21.86% | **21.81%** | 22.26% | 23.68% |
| fr-es | **21.30%** | 21.47% | 22.33% | 23.70% |
| it-es | **21.69%** | 21.93% | 22.93% | 24.57% |
| pt-es | **20.87%** | 21.07% | 21.78% | 23.00% |
| fr-en | 22.37% | **22.19%** | 22.83% | 24.36% |
| es-ca | **19.22%** | 19.67% | 19.70% | 19.87% |
| en-fi | 24.14% | **24.01%** | 25.91% | 27.29% |
| cy-en | 22.30% | 21.86% | **21.39%** | 21.98% |

**Table 12:** Compression ratios obtained with the searchable 2LCAB compressor.

We have studied how the time needed to process a query depends on the language pair and also on the number and frequencies of the words in the query. The average times over 100 different sequences and 10 runs are reported in Figures 4 and 5, in which process times were measured on an AMD Athlon Dual Core at 2 GHz with 2GB of RAM.

Figure 4 presents the times for two different language pairs. The first one, `en-fr`, displays the typical behavior of all Europarl bitexts (Koehn, 2005), while the second one, `en-fi`, requires particularly longer times, especially for large queries. This divergent behavior seems to originate in the poor quality of the alignments between the words in this pair of languages. This often makes words participate in a large number of different biwords and this degrades the performance of the SET-HORSPOOL algorithm. This language pair consistently leads to the worst compression ratios.

Finally, Figure 5 shows the processing times for two language pairs (`en-cy` and `es-ca`) whose bitexts have been obtained from a totally different source. The processing times are considerably lower than those required by the Europarl corpus and a manual inspection of the bitexts revealed that they have a highly parallel structure. This implies that the words participate only in a small number of biwords and, not surprisingly, 2LCAB achieves the lowest compression ratios with these language pairs.

## 7. Concluding Remarks

We have introduced the concept of generalized biwords when applied to the compression of bitexts. Generalized biwords integrate the information concerning word reordering and multiword expressions in the translated text. We have described a procedure that transforms the bitext into a sequence of generalized biwords which can be used as an intermediate representation in the compression process. We have then extended the binary compression algorithm 2LCAB and proposed a new one, called TRE, for the encoding of generalized biword sequences. We have also designed a variant of the 2LCAB compression technique, and a companion algorithm which facilitates efficient searching and translation spotting on the compressed bitext.

The compression performance of 2LCAB and TRE has been tested with four different schemes to extract the biword sequence, each of which uses biwords with different structural

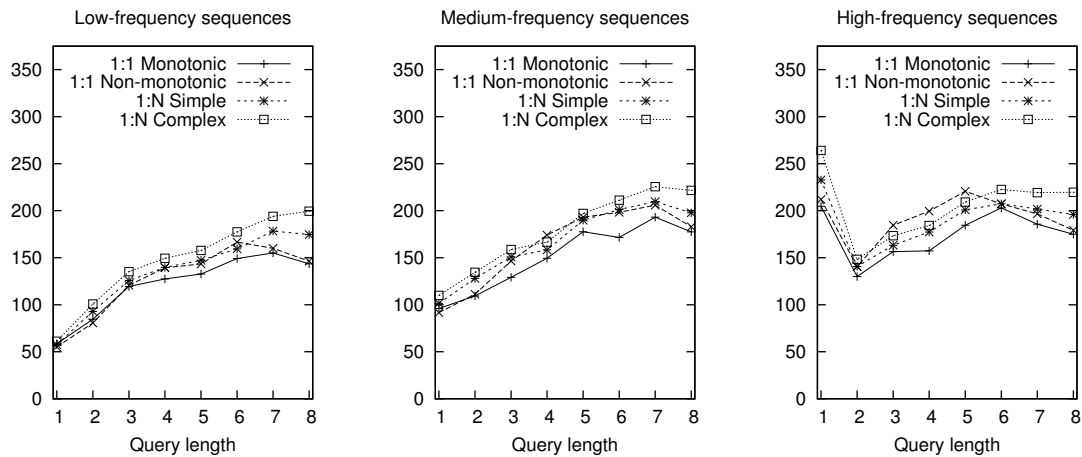

(a) English–French (`en-fr`) bitext

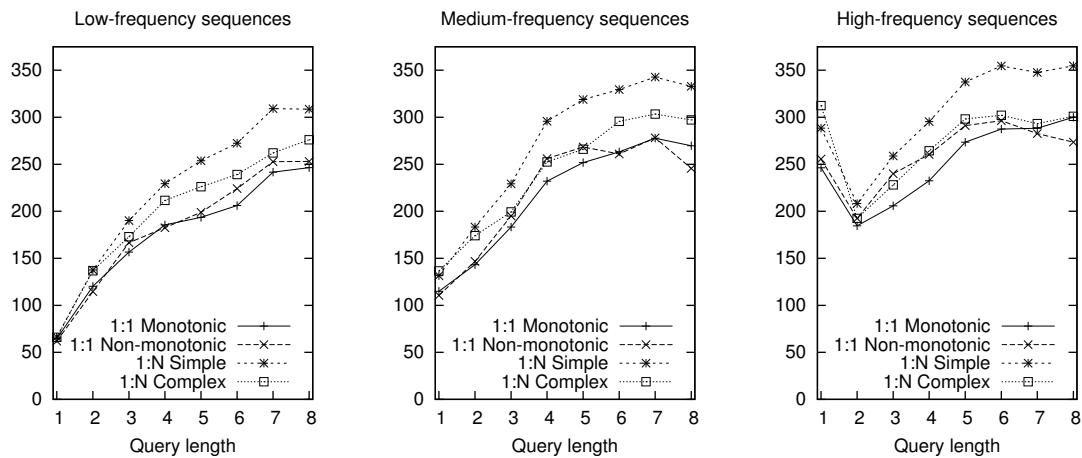

(b) English–Finnish (`en-fi`) bitext

**Figure 4:** Average time (milliseconds) needed to process a query containing only words with low, medium or high frequency, as a function of the query length. Times are shown for two different language pairs and four encoding methods.

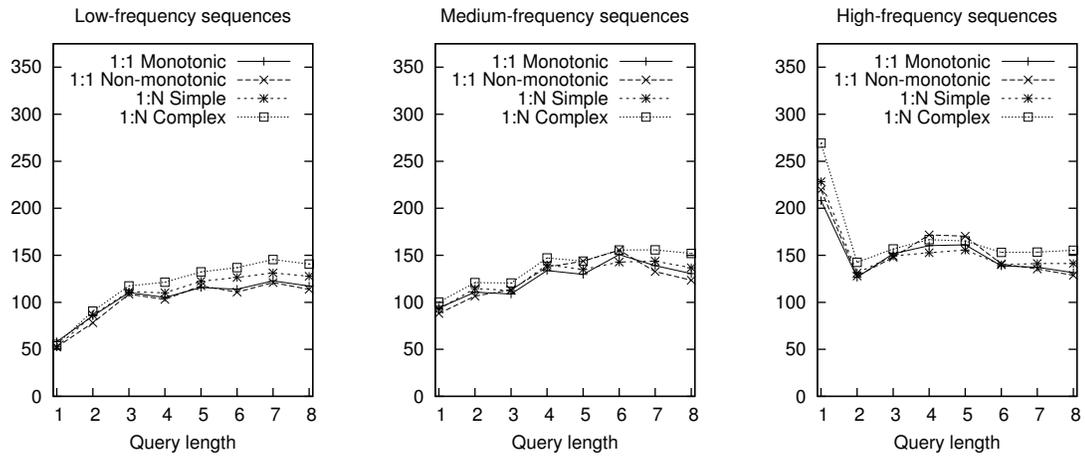

(a) English–Welsh (`en-cy`) bitext

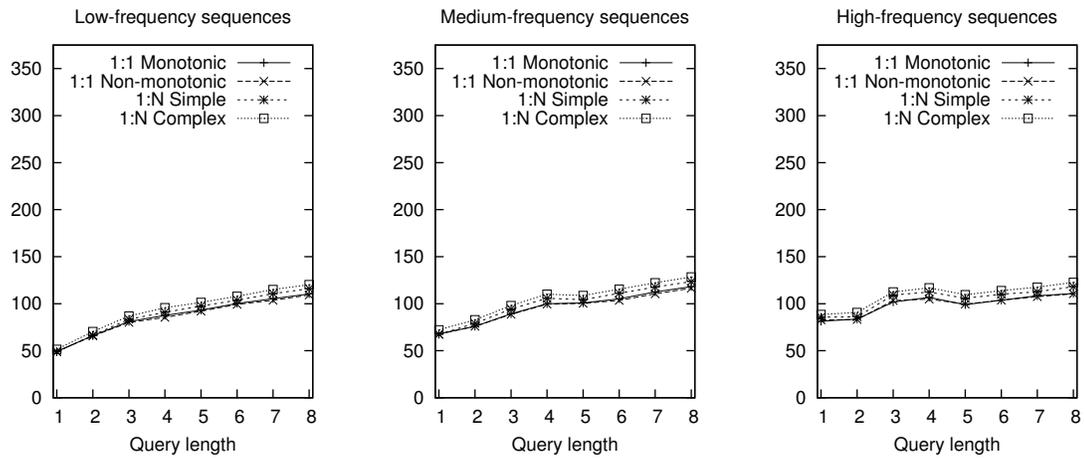

(b) Spanish–Catalan (`es-ca`) bitext

**Figure 5:** Average time (milliseconds) needed to process a query for two different language pairs (`en-cy` and `es-ca`).

complexities. The simplest method uses biwords without shifts and is therefore equivalent to those approaches in which biwords are simple pairs and include no structural information. The other methods include offsets to integrate the structural information of the alignments.

Our experiments show that generalized biwords lead to better compression ratios because the reduction in the sequences encoding the bitext compensates the larger dictionaries needed. The largest reduction in the compression ratios is obtained for pairs of divergent languages because, in these cases, biwords without shifts cannot tackle the frequent word reorderings and multiword translations.

Since the enhanced variability of generalized biwords requires larger dictionaries that increase the header included in the compressed files, we have tested the effect of splitting infrequent biwords into smaller, more frequent biwords. This reduces the number of different biwords and allows the 2LCAB compressor to obtain lower compression ratios. After this pruning, 2LCAB provides the best results if biwords are obtained from one-to-many word alignments in which only simple shifts are allowed, that is, the target text is only split into segments of contiguous words. Both of the algorithms, TRE and 2LCAB, provide better compression ratios than general purpose compressors, particularly in the case of pairs of languages that share a common language family (`es-ca`) or bitexts which are highly parallel (`en-cy`). The compression ratio can therefore be used to indirectly measure the quality of word alignment and the degree of parallelism of the bitext.

Some modifications made to the 2LCAB compressors allow the compressed bitext to be searched efficiently, although this adaptation leads to slightly worse compression ratios. However, the new compressed file includes the alignments between the words in the bitext and this additional information is needed in order to implement translation spotting.

The relatively small difference in the time needed to process a query by the `1:1 Monotonic` method (the fastest one) and by the `1:N Complex` method (the one conveying most information) makes the latter the preferable choice in translation spotting because it identifies a larger variety of translations in the bitext and provides richer examples.

In our future work we plan to study the effect on translation performance of the integration of generalized, biword-based bilingual language models into current state-of-the-art statistical machine translation systems.

## Appendix A. Biword Extraction Algorithm

Algorithm 1 shows the procedure used to obtain the sequence of generalized biwords $\mathcal{B}$ from a bitext with one-to-many word alignments. The main loop (lines 3–25) iterates over the words in the left and right texts while there are still words on both sides to be considered. Variables $m$ and $n$ point to the next left and right words, respectively, to be processed. Inside the main loop, the set $A_m^{\rightarrow}$ with the positions of the right words aligned with the left word $l_m$, and the set $A_n^{\leftarrow}$ with the positions of the left words aligned with the right word $r_n$ are first computed. As word alignments are one-to-many, $A_n^{\leftarrow}$ contains, at most, one element.

After every iteration, a single biword is produced. If $A_m^{\rightarrow}$ is empty, i.e., $l_m$ is not aligned, the next biword consists of the left word $l_m$, an empty sequence of right words and an empty sequence of offsets (line 7). If $A_m^{\rightarrow}$ is not empty but $A_n^{\leftarrow}$ is, the biword consists of the empty word, a sequence of right words containing only $r_n$ and a sequence of offsets containing only

---

**Algorithm 1** GETBIWORDS extracts a biword sequence from a one-to-many word alignment between two texts.

---

**Input:** Two word sequences $L$ and $R$, and a one-to-many bigraph $G = \{L, R, A\}$

**Output:** A sequence $\mathcal{B}$ of 3-tuples (word, sequence of words, sequence of offsets)

1:   $\mathcal{B} \leftarrow ()$              $\triangleright$ Create an empty sequence of 3-tuples

2:   $m \leftarrow 1; n \leftarrow 1$

3:   **while** $(m \leq M) \wedge (n \leq N)$ **do**

4:       $A_m^{\rightarrow} \leftarrow \{j : (l_m, r_j) \in A\}$

5:       $A_n^{\leftarrow} \leftarrow \{i : (l_i, r_n) \in A\}$

6:       **if** $A_m^{\rightarrow} = \emptyset$ **then**

7:          **add** $(l_m, (), ())$ **to** $\mathcal{B}$

8:          $m \leftarrow m + 1$

9:       **else if** $A_n^{\leftarrow} = \emptyset$ **then**

10:         **add** $(\epsilon, (r_n), (0))$ **to** $\mathcal{B}$

11:         $n \leftarrow$ NEXTRIGHT$(m, n, G)$

12:       **else**

13:         $\rho \leftarrow (); \omega \leftarrow ()$          $\triangleright$ Create empty sequences of words and offsets

14:         $k \leftarrow n$

15:         **for all** $j \in A_m^{\rightarrow}$ **in** ascending order **do**

16:            **add** $r_j$ **to** $\rho$; **add** $j - k$ **to** $\omega$

17:            $k \leftarrow j + 1$

18:            **if** $n = j$ **then**

19:               $n \leftarrow$ NEXTRIGHT$(m, n, A)$

20:            **end if**

21:         **end for**

22:         **add** $(l_m, \rho, \omega)$ **to** $\mathcal{B}$

23:         $m \leftarrow m + 1$

24:       **end if**

25:   **end while**

26:   **while** $m \leq M$ **do**

27:       **add** $(l_m, (), ())$ **to** $\mathcal{B}$

28:       $m \leftarrow m + 1$

29:   **end while**

30:   **while** $n \leq N$ **do**

31:       $A_n^{\leftarrow} \leftarrow \{i : (l_i, r_n) \in A\}$

32:       **if** $A_n^{\leftarrow} = \emptyset$ **then**

33:         **add** $(\epsilon, (r_n), (0))$ **to** $\mathcal{B}$

34:       **end if**

35:       $n \leftarrow n + 1$

36:   **end while**

37:   **return** $\mathcal{B}$

---

---

**Algorithm 2** NextRight

---

**Input:** Integers $m$ and $n$, one-to-many bigraph $G = (L, R, A)$
**Output:** Index of the next right word paired with $\epsilon$, with $l_m$ or posterior word in $L$
1: **repeat**
2:    $n \leftarrow n + 1$
3:    $A_n^{\leftarrow} \leftarrow \{i : (l_i, r_n) \in A\}$
4: **until** $(n > N) \vee (A_n^{\leftarrow} = \emptyset) \vee (\min(A_n^{\leftarrow}) \geq m)$
5: **return** n

---

a null offset (line 10). Otherwise, the biword consists of $l_m$, a sequence containing all the right words aligned with $l_m$, and a sequence containing one offset for each right word. The first offset is relative to $n$, whereas the following ones are relative to the previous word in the sequence (see lines 15–21).

Index $m$ is simply incremented every time a biword containing a left word is produced. The update of $n$ is more subtle since some words in positions greater than $n$ may have already been processed because they are aligned with a left word preceding $l_m$. $n$ is therefore assigned the value returned by function NextRight (depicted in Algorithm 2) which, given the current values of $m$ and $n$, looks for the next $n$ such that $r_m$ is paired either with the empty word $\epsilon$ or with a left word not preceding $l_m$.

Finally, two loops take care of the words that remain unprocessed after the main loop.

## Appendix B. Bitext Restoration Algorithm

Algorithm 3 provides the pseudo-code with which to restore the right text of the bitext from the biword representation obtained with Algorithm 1. Restoring the left text is straightforward since biwords are sorted by their left component.

The main loop in Algorithm 3 (lines 3–15) iterates over the sequence of biwords representing the bitext. Variables $m$ and $n$ point to the next biword to be processed, and to the next gap in $R$ to be filled in with a word, respectively. It then iterates over the array of offsets $\omega = (w_1, \ldots, w_{|\omega|})$ (lines 7–10) and places each word $\rho_j$ in the sequence of right words $\rho = (\rho_1, \ldots, \rho_{|\omega|})$ in the right place. After each biword has been processed, $m$ is updated to point to the next biword, and $n$ to point to the next gap in $R$ to be filled in (lines 12–14).

## Acknowledgments

This work has been supported by the Spanish Government through projects TIN2009-14009-C02-01 and TIN2009-14009-C02-02, and by the Millennium Institute for Cell Dynamics and Biotechnology (grant ICM P05-001-F). During the development of the work reported in this paper, Miguel A. Martínez-Prieto was at the Department of Computer Science (University of Chile) on a post-doctoral stay. The authors thanks Nieves R. Brisaboa for her ideas and cooperation in the development of the initial version of 2lcab, Gonzalo Navarro for his inspiration for the Tre compression approach, the anonymous referees for suggesting significant improvements to this paper and Francis M. Tyers for proof-reading it.

---

**Algorithm 3** GetRightText retrieves the right text from the biword representation of the bitext.

---

**Input:** A sequence of biwords $\mathcal{B} = (\beta_1, \ldots, \beta_M)$
**Output:** The right text $R = r_1 r_2 \cdots r_N$ contained in the sequence $\mathcal{B}$.

1: $m \leftarrow 1$
2: $n \leftarrow 1$
3: **while** $m \leq M$ **do**
4:    $k = n - 1$
5:    $\omega \leftarrow \mathtt{offset}(\beta_m)$
6:    $\rho \leftarrow \mathtt{right}(\beta_m)$
7:    **for** $j = 1, \ldots, |\omega|$ **do**
8:      $k \leftarrow k + \omega_j + 1$
9:      $r_k \leftarrow \rho_j$
10:    **end for**
11:    $m \leftarrow m + 1$
12:    **while** $r_n$ is not undefined **do**
13:      $n \leftarrow n + 1$
14:    **end while**
15: **end while**
16: **return** $R$

---

# References

Adiego, J., Brisaboa, N. R., Martínez-Prieto, M. A., & Sánchez-Martínez, F. (2009). A two-level structure for compressing aligned bitexts. In *Proceedings of the 16th String Processing and Information Retrieval Symposium*, Vol. 5721 of *Lecture Notes in Computer Science*, pp. 114–121, Saariselkä, Finland. Springer.

Adiego, J., & de la Fuente, P. (2006). Mapping words into codewords on PPM. In *Proceedings of the 13th String Processing and Information Retrieval Symposium*, Vol. 4209 of *Lecture Notes in Computer Science*, pp. 181–192, Glasgow, UK. Springer.

Adiego, J., Martínez-Prieto, M. A., Hoyos-Torio, J. E., & Sánchez-Martínez, F. (2010). Modelling parallel texts for boosting compression. In *Proceedings of the 2010 Data Compression Conference*, p. 517, Snowbird, USA.

Arnold, D., Balkan, L., Meijer, S., Humphreys, R., & Sadler, L. (1994). *Machine translation: An introductory guide*. NCC Blackwell, Oxford.

Barlow, M. (2004). Parallel concordancing and translation. In *Proceedings of ASLIB Translating and the Computer 26*, London, UK.

Bell, T. C., Cleary, J. G., & Witten, I. H. (1990). *Text compression*. Prentice Hall.

Bourdaillet, J., Huet, S., Langlais, P., & Lapalme, G. (2010). TransSearch: from a bilingual concordancer to a translation finder. *Machine Translation, 23*(3–4), 241–271. Published in 2011.

Bowker, L., & Barlow, M. (2004). Bilingual concordancers and translation memories: a comparative evaluation. In *Proceedings of the Second International Workshop on*

*Language Resources for Translation Work, Research and Training at Coling 2004*, pp. 70–79, Geneva, Switzerland.

Boyer, R., & Moore, J. S. (1977). A fast string searching algorithm. *Communications of the ACM*, *20*(10), 762–772.

Brisaboa, N., Ladra, S., & Navarro, G. (2009). Directly addressable variable-length codes. In *Proceedings of the 16th String Processing and Information Retrieval Symposium*, Vol. 5721 of *Lecture Notes in Computer Science*, pp. 122–130, Saariselkä, Finland. Springer.

Brisaboa, N. R., Fariña, A., Navarro, G., & Paramá, J. R. (2007). Lightweight natural language text compression. *Information Retrieval*, *10*(1), 1–33.

Brown, P., Lai, J., & Mercer, R. (1991). Aligning sentences in parallel corpora. In *Proceedings of the 29th Annual Meeting of the Association for Computational Linguistics*, pp. 169–176, Berkeley, CA, USA.

Brown, P. F., Cocke, J., Pietra, S. A. D., Pietra, V. J. D., Jelinek, F., Lafferty, J. D., Mercer, R. L., & Roossin, P. S. (1990). A statistical approach to machine translation. *Computational Linguistics*, *16*(2), 76–85.

Brown, P. F., Pietra, S. A. D., Pietra, V. J. D., & Mercer, R. L. (1993). The mathematics of statistical machine translation: Parameter estimation. *Computational Linguistics*, *19*(2), 263–311.

Burrows, M., & Wheeler, D. (1994). A block sorting lossless data compression algorithm. Tech. rep. 124, Digital Systems Research Center, Palo Alto, CA, USA.

Callison-Burch, C., Bannard, C., & Schroeder, J. (2005a). A compact data structure for searchable translation memories. In *Proceedings of the 10th European Association for Machine Translation Conference*, pp. 59–65, Budapest, Hungary.

Callison-Burch, C., Bannard, C., & Schroeder, J. (2005b). Scaling phrase-based statistical machine translation to larger corpora and longer phrases. In *Proceedings of the 43rd Annual Meeting of the Association for Computational Linguistics*, pp. 255–262, Ann Arbor, USA.

Carl, M., & Way, A. (Eds.). (2003). *Recent Advances in Example-Based Machine Translation*, Vol. 21 of *Text, Speech and Language Technology*. Springer.

Carroll, J. (2003). *The Oxford Handbook of Computational Linguistics*, chap. 12 Parsing, pp. 233–248. Oxford University Press.

Casacuberta, F., & Vidal, E. (2004). Machine translation with inferred stochastic finite-state transducers. *Computational Linguistics*, *30*(2), 205–225.

Clark, D. (1996). *Compact PAT trees*. Ph.D. thesis, University of Waterloo, Warteloo, ON, Canada.

Cleary, J. G., & Witten, I. H. (1984). Data compression using adaptive coding and partial string matching. *IEEE Transactions on Communications*, *32*(4), 396–402.

Conley, E., & Klein, S. (2008). Using alignment for multilingual text compression. *International Journal of Foundations of Computer Science*, *19*(1), 89–101.

Cover, T. M., & Thomas, J. A. (1991). *Elements of Information Theory*. Wiley.

Gale, W. A., & Church, K. W. (1993). A program for aligning sentences in bilingual corpora. *Computational Linguistics*, *19*(1), 75–102.

González, R., Grabowski, S., Mäkinen, V., & Navarro, G. (2005). Practical implementation of rank and select queries. In *Proceedings of the 4th International Workshop on Efficient and Experimental Algorithms*, pp. 27–38, Santorini Island, Greece.

Grossman, D. A., & Frieder, O. (2004). *Information Retrieval: Algorithms and Heuristics* (2nd edition)., Vol. 15 of *The Information Retrieval Series*. Springer.

Hasan, S., Ganitkevitch, J., Ney, H., & Andrés-Ferrer, J. (2008). Triplet lexicon models for statistical machine translation. In *Proceedings of the 2008 Conference on Empirical Methods on Natural Language Processing*, pp. 372–381, Honolulu, USA.

Horspool, R. N. (1980). Practical fast searching in strings. *Software: Practice and Experience*, *10*(6), 501–506.

Howard, P., & Vitter, J. (1992). Practical implementations of arithmetic coding. In Storer, J. (Ed.), *Image and Text Compression*, pp. 85–112. Kluwer Academic.

Huffman, D. (1952). A method for the construction of minimum-redundancy codes. *Proceedings of the Institute of Radio Engineers*, *40*(9), 1098–1101.

Ide, N., & Véronis, J. (1998). Word sense disambiguation: The state of the art. *Computational Linguistics*, *24*(1), 1–41.

Jones, D., & Eisele, A. (2006). Phrase-based statistical machine translation between English and Welsh. In *Proceedings of the 5th SALTMIL Workshop on Minority Languages at the 5th International Conference on Language Resources and Evaluation*, pp. 75–77, Genoa, Italy.

Knuth, D. E., Morris, J. H., & Pratt, V. (1977). Fast pattern matching in strings. *SIAM Journal on Computing*, *6*(2), 323–350.

Koehn, P. (2005). Europarl: A parallel corpus for statistical machine translation. In *Proceedings of the Tenth Machine Translation Summit*, pp. 79–86, Phuket, Thailand.

Koehn, P. (2010). *Statistical Machine Translation*. Cambridge University Press.

Lopez, A. (2007). Hierarchical phrase-based translation with suffix arrays. In *Proceedings of the 2007 Joint Conference on Empirical Methods in Natural Language Processing and Computational Natural Language Learning*, pp. 976–985, Prague, Czech Republic.

Lopez, A. (2008). Statistical machine translation. *ACM Computing Surveys*, *40*(3), 1–49.

Manber, U., & Myers, G. (1993). Suffix arrays: a new method for on-line string searches. *SIAM Journal on Computing*, *22*(5), 935–948.

Manning, C. D., & Schütze, H. (1999). *Foundations of statistical natural language processing*. MIT Press.

Mariño, J., Banchs, R. E., Crego, J. M., de Gispert, A., Lambert, P., Fonollosa, J. A. R., & Costa-Jussà, M. R. (2006). N-gram-based machine translation. *Computational Linguistics*, *32*(4), 527–549.

Martin, G. (1979). Range encoding: an algorithm for removing redundancy from a digitized message. In *Proceedings of Video and Data Recording Conference*, Southampton, UK.

Martínez-Prieto, M. A., Adiego, J., Sánchez-Martínez, F., de la Fuente, P., & Carrasco, R. C. (2009). On the use of word alignments to enhance bitext compression. In *Proceedings of the 2009 Data Compression Conference*, p. 459, Snowbird, USA.

Matusov, E., Zens, R., Vilar, D., Mauser, A., Popović, M., Hasan, S., & Ney, H. (2006). The RWTH machine translation system. In *Proceedings of the TC-STAR Workshop on Speech-to-Speech Translation*, pp. 31–36, Barcelona, Spain.

Melamed, I. D. (2001). *Emplirical methods for exploting parallel texts*. MIT Press.

Mihalcea, R., & Simard, M. (2005). Parallel texts. *Natural Language Engineering, 11*(3), 239–246.

Moffat, A. (1989). Word-based text compression. *Software: Practice and Experience, 19*(2), 185–198.

Moffat, A., & Isal, R. Y. K. (2005). Word-based text compression using the Burrows-Wheeler transform. *Information Processing and Management, 41*(5), 1175–1192.

Moffat, A., & Turpin, A. (1997). On the implementation of minimum redundancy prefix codes. *IEEE Transactions on Communications, 45*(10), 1200–1207.

Navarro, G., & Mäkinen, V. (2007). Compressed full-text indexes. *ACM Computing Surveys, 39*(1). Article 2.

Navarro, G., & Raffinot, M. (2002). *Flexible Pattern Matching in String: Practical on-line search algorithms for texts and biological sequences*. Cambridge University Press.

Nevill-Manning, C., & Bell, T. (1992). Compression of parallel texts. *Information Processing & Management, 28*(6), 781–794.

Niehues, J., Herrmann, T., Vogel, S., & Waibel, A. (2011). Wider context by using bilingual language models in machine translation. In *Proceedings of the 6th Workshop on Statistical Machine Translation*, pp. 198–206, Edinburgh, UK.

Och, F. J., & Ney, H. (2003). A systematic comparison of various statistical alignment models. *Computational Linguistics, 29*(1), 19–51.

Raman, R., Raman, V., & Rao, S. (2002). Succinct indexable dictionaries with applications to encoding *k*-ary trees and multisets. In *Proceedings of the 22nd Annual ACM-SIAM Symposium on Discrete Algorithms*, pp. 233–242, San Francisco, CA, USA.

Salomon, D. (2007). *Data compression. The complete reference* (Fourth edition). Springer.

Shkarin, D. (2002). PPM: One step to practicality. In *Proceeding of the 2002 Data Compression Conference*, pp. 202–211, Snowbird, USA.

Simard, M. (2003). Translation spotting for translation memories. In *Proceedings of NAACL 2003 Workshop on Building and Using Parallel Texts: Data Driven Machine Translation and Beyond*, pp. 65–72, Edmonton, AB, Canada.

Suel, T., & Memon, N. (2003). Algorithms for delta compression and remote file synchronization. In Sayood, K. (Ed.), *Lossless Compression Handbook*, pp. 269–290. Academic Press.

Tufis, D., Barbu, A.-M., & Ion, R. (2004). Extracting multilingual lexicons from parallel corpora. *Computers and the Humanities*, *38*(2), 163–189.

Turpin, A., & Moffat, A. (2000). Housekeeping for prefix coding. *IEEE Transactions on Communications*, *48*(4), 622–628.

Véronis, J., & Langlais, P. (2000). *Parallel text processing. Alignment and use of translation corpora*, chap. Evaluation of Parallel Text Alignment Systems – The ARCADE Project. Kluwer Academic Publishers.

Vogel, S., Ney, H., & Tillmann, C. (1996). HMM-based word alignment in statistical translation. In *Proceedings of the 16th International Conference on Computational Linguistics*, pp. 836–841, Copenhagen, Denmark.

Ziv, J., & Lempel, A. (1977). An universal algorithm for sequential data compression. *IEEE Transactions on Information Theory*, *23*(3), 337–343.

Ziviani, N., Moura, E., Navarro, G., & Baeza-Yates, R. (2000). Compression: A key for next-generation text retrieval systems. *IEEE Computer*, *33*(11), 37–44.